\documentclass[letterpaper]{article} 
\usepackage[preprint]{aaai2027} 
\usepackage[hyphens]{url} 
\usepackage{graphicx} 
\usepackage{natbib} 
\usepackage{caption} 
\usepackage{amsmath}
\usepackage{booktabs}
\usepackage{multirow}
\usepackage{subcaption}
\AtBeginDocument{}
\begin{document}

\title{Therapy-FIS: A Dataset for Assessing Facilitative Interpersonal Skills in Psychotherapeutic Setting}

\author{
Qijian Zheng\textsuperscript{1},
Wanqing Lin\textsuperscript{2},
Yuanfang Wang\textsuperscript{3},
Yijun Ge\textsuperscript{4}\\
Jordan Bate\textsuperscript{5},
Katie Aafjes-van Doorn\textsuperscript{2}\corresponding,
Feng Liu\textsuperscript{4}\corresponding
}
\affiliations{
\textsuperscript{1}College of Computer Science and Artificial Intelligence, Fudan University, Shanghai, P. R. China\\
\textsuperscript{2}Faculty of Arts and Sciences, New York University Shanghai, Shanghai, P. R. China\\
\textsuperscript{3}Global College, Shanghai Jiao Tong University, Shanghai, P. R. China\\
\textsuperscript{4}School of Psychology, Shanghai Jiao Tong University, Shanghai, P. R. China\\
\textsuperscript{5}Adelphi University, New York, USA\\
qjzheng25@m.fudan.edu.cn, wl3215@nyu.edu\\
zhaomc911@sjtu.edu.cn, yijun.ge@sjtu.edu.cn, jbate@adelphi.edu\\
kav9239@nyu.edu, liu.feng@sjtu.edu.cn
}

\maketitle

\begin{abstract}
Facilitative Interpersonal Skills (FIS) is a performance-based assessment method that provides a structured framework for evaluating the interpersonal competencies of mental health professionals. However, existing FIS assessment relies heavily on human observer ratings, making the evaluation process labor-intensive, costly, and difficult to scale. Progress toward automated FIS assessment has also been constrained by the lack of publicly available professional data in psychotherapy settings. To address this limitation, we introduce Therapy-FIS, a dataset for automated FIS assessment in standardized psychotherapy tasks. To our knowledge, Therapy-FIS is the first publicly available dataset in this domain. It contains facial, acoustic, and textual signals derived from standardized therapist responses to standardized mock patient stimulus clips, based on 890 therapist video recordings and 8 patient videos, and is rated with eight FIS subscale ratings and an overall FIS score. Built on Therapy-FIS, we further establish a benchmark with two downstream tasks: Therapist Performance Rating and Joint Therapist--mock-Patient Performance Rating. We also introduce a competitive baseline, DyadicFIS, with gated fusion and routed readout under therapist-only and dyadic settings. DyadicFIS obtains the best macro CCC(Concordance Correlation Coefficient) scores on both tasks ($0.2807$ and $0.3299$). Therapy-FIS offers a reproducible resource for data-driven automated assessment of therapeutic interpersonal skills in mental health.
\end{abstract}

\begin{figure*}[t]
  \includegraphics[width=\textwidth]{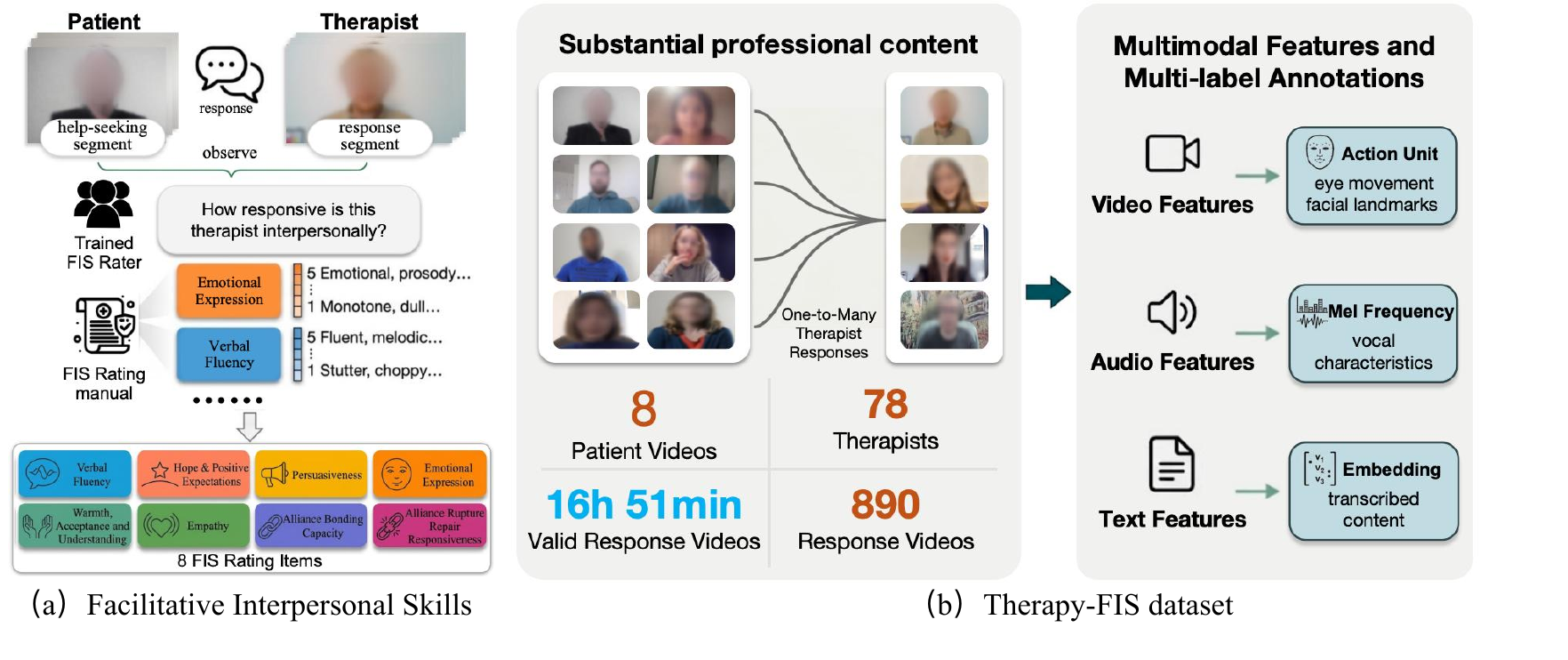}
  \caption{Overview of the Therapy-FIS dataset and the FIS evaluation protocol. (a)~Facilitative interpersonal skills are evaluated by observing standardized mock patient stimulus clips and the therapist's response segment, and then rating the therapist on eight FIS items. (b)~Therapy-FIS releases derived behavioral features from therapist responses and mock-patient stimuli under the FIS protocol.}
  \label{fig:1}
\end{figure*}


\section{Introduction}

Facilitative Interpersonal Skills (FIS) have been established as robust predictors of the therapeutic alliance and patient outcomes \cite{allen_review_2024, anderson_therapist_2009, anderson_therapist_2016}. Research demonstrates that a substantial portion of the variance in psychotherapy success stems from therapist effects, which operate independently of the specific treatment modality and are largely explained by FIS \cite{schiefeleReliabilityTherapistEffects2017, johnsSystematicReviewTherapist2019}. Instead of measuring general self-perceived competence, the assessment of FIS targets the observable interpersonal qualities exhibited by therapists responding to mock patient stimulus clips\cite{anderson_facilitative_2020}. Therapists exhibiting higher FIS—defined as the ability to respond to patients in ways that are empathic, emotionally attuned, and therapeutically constructive—facilitate greater symptom reduction \cite{ aafjes-van_doorn_will_2024,  de_jong_facilitative_2024}. In this sense, FIS provides a structured concept for characterizing therapy skill at the therapist level and offers a behavior-grounded target for computational assessment. 

Existing FIS assessment remains dependent on trained human observer raters, which makes it labor-intensive and hard to scale in large-scale clinical training \cite{imelComputationalPsychotherapyResearch2015}. Early computational efforts initially explored unimodal text \cite{goldberg_can_2021} and audio signals \cite{glasgow_vocal_2023}. Broader research on automatic analysis of therapy-like interactions has also explored combining complementary behavioral cues \cite{ zhouan_zhu_medic_2023, kashihara_constructing_2024, review_neuarl_model}, yet progress toward automated FIS assessment remains constrained by the absence of publicly available professional data under the standardized FIS protocol. This gap is partly due to the demanding requirements of FIS data collection: standardized stimulus design, recruitment of trained or in-training mental health professionals and reliable annotation by trained human observer raters, and navigating stringent privacy and ethical constraints inherent in clinical interactions \cite{cumminsReviewDepressionSuicide2015}.

To address this gap, we introduce Therapy-FIS, a therapy-centered dataset for FIS assessment. Therapy-FIS organizes therapist response recordings together with mock patient-side stimulus information under the standardized FIS framework, and releases de-identified derived behavioral signals such as facial action units, acoustic cues, and textual content to facilitate reproducible research while respecting privacy constraints. Based on this resource, we further formulate automatic FIS evaluation tasks in therapist-only and therapist--mock-patient settings and provide benchmark experiments with representative baselines.
Our contributions are threefold:
\begin{enumerate}
    \item \textbf{A therapy-centered FIS dataset.} We introduce Therapy-FIS, built from standardized FIS experiments that organizes video-recorded therapist responses, mock patient-side stimuli, and de-identified derived signals including facial action units, acoustic cues, and textual content.
    \item \textbf{A benchmark on FIS evaluation.} We establish a benchmark on Therapy-FIS for automatic FIS evaluation by formulating prediction tasks from both therapist-only and therapist--mock-patient joint perspectives. These tasks operationalize therapy skill assessment within a computational framework, and our reported results provide a reproducible starting point for scalable automatic evaluation of FIS.
    \item \textbf{A competitive baseline for the proposed protocol.} We introduce DyadicFIS as a baseline under the Therapy-FIS tasks. Experiments show that DyadicFIS achieves strong agreement with human FIS ratings under the proposed evaluation protocol.
    
\end{enumerate}

\section{Related Work}

\subsection{Facilitative Interpersonal Skills Dataset}
The FIS performance-based task and rating method systematically evaluates therapist competencies through standardized, interpersonally challenging clinical vignettes \cite{anderson_therapist_2009}. Extensive research confirms its robust reliability and construct validity, with scores consistently predicting therapeutic alliance and patient symptom reduction more effectively than self-reported metrics \cite{ anderson2019facilitative, aafjes-van_doorn_will_2024, schottkePredictingPsychotherapyOutcome2017}. Consequently, the FIS framework has been adapted for diverse contexts, including child therapy \cite{bateFacilitativeInterpersonalSkills2022}, text-based interventions \cite{zechAutomaticRatingTherapist2022}, and remote environments \cite{lin_efficacy_2025, antebi-lermanAssessingTherapistSkills2026}.

While computational psychology has introduced datasets for analyzing therapeutic interactions, most focus on general conversational dynamics or specific constructs like empathy or alliance \cite{ tahirETHERMultimodalDataset2025, zhangYouNeverKnow2023, linCOMPASSComputationalMapping2025}. However, naturalistic session data often introduces excessive variability in patient pathology, making it ill-suited for the standardized evaluation of performance-based competencies \cite{ fairburnTherapistCompetenceTherapy2011, imelComputationalPsychotherapyResearch2015}. Therapy-FIS addresses this critical gap by providing a large-scale, publicly available dataset specifically designed for the standardized assessment of therapeutic skills.

\subsection{Interpersonal Skills Evaluation}
The FIS performance-based task and rating method comprises eight subscales that operationalize general therapeutic skills, unrelated to treatment approach or patient population \cite{ anderson2019facilitative}. Extensive empirical evidence, including randomized trials, demonstrates that proficiency in subscales such as Empathy, Warmth, and Alliance Rupture-Repair Responsiveness predicts treatment success and symptom reduction more effectively than therapist demographics or self-reported traits \cite{ wampold_great_2015, fluckigerAllianceAdultPsychotherapy2018, eubanksAllianceRuptureRepair2018, elliottEmpathy2011, abargil_how_2022,  heinonenProfessionalPersonalCharacteristics2020}. While human observer rating remains the traditional standard, the labor-intensive nature of manual rating necessitates scalable, automated assessment methods \cite{anderson_facilitative_2020}.

Early computational approaches focused on unimodal analysis, utilizing machine learning and NLP to model clinical constructs from session transcripts, acoustic features, or non-verbal synchrony \cite{goldberg_machine_2020, ryu_natural_2023, eberhardt_decoding_2025, lalk_employing_2025, tanana_how_2021, cohen_nonverbal_2021, reinecke_hand_2022}. However, reliance on a single data stream can miss complementary behavioral evidence and limit predictive power \cite{hinduja_multimodal_2024}. Recent work therefore combines linguistic, acoustic, and visual cues for automatic estimation of therapy-related constructs, including FIS ratings \cite{tavabi_multimodal_2020, aafjes-van_doorn_development_2025}. Our focus is the standardized FIS evaluation scenario and a reproducible public benchmark for scalable skill assessment in psychotherapeutic settings.

\section{FIS Dataset}
\subsection{Experimental Setup}
\begin{figure*}
    \centering
    \includegraphics[width=\linewidth]{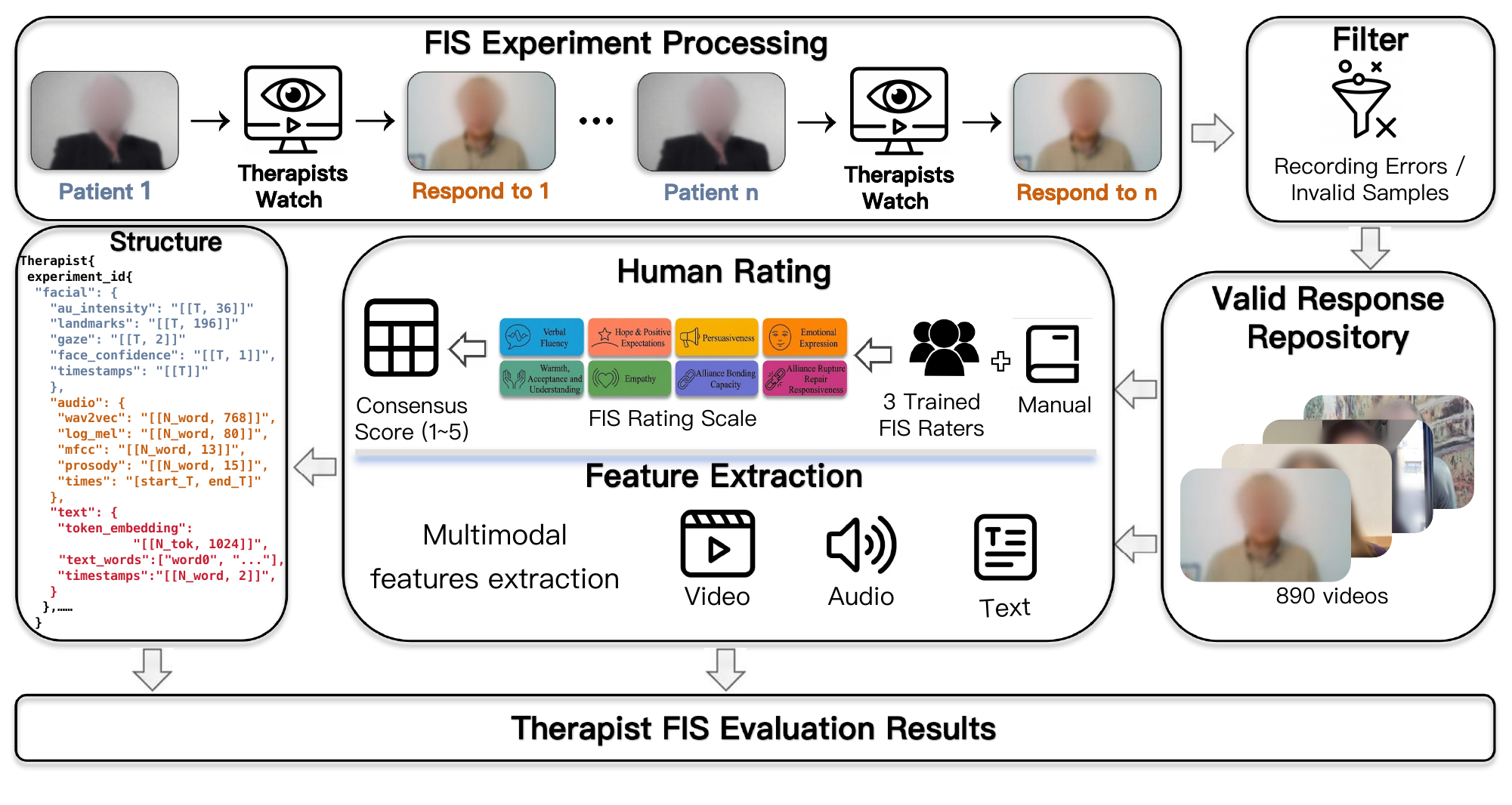}
    \caption{Dataset construction pipeline overview. FIS experiment: therapists watch a patient’s help-seeking segment and provide a recorded response. Human ratings: each response clip is rated by three trained FIS raters, and the final score is the average of the three ratings, and the overall FIS score is the mean of the eight averaged subscales. Structure: each therapist’s derived behavioral features are stored in tensor files.}
    \label{fig:fis_statistics}
\end{figure*}
\textbf{Equipment.} The video was recorded via Zoom and stored at a resolution of 1920×1080 pixels with a frame rate of 30 fps. Audio was captured using the recording device’s built-in microphone and stored together with the video stream. 

\textbf{Standardized psychotherapy scenarios.} The data collection followed a standardized FIS setting. Eight actors served as mock patients representing different challenging clinical scenarios typically encountered in psychotherapy, and they enacted pre-defined psychotherapy sessions. Based on patient scripts and transcripts, these scenarios were designed to elicit responses to clinically challenging situations, such as self-blame, confrontation, professional dilemmas and doubts about treatment effectiveness. This design enabled participating mental health professionals to respond to comparable yet ecologically meaningful psychotherapy situations. The response clips analyzed in the present study were drawn from a large-scale FIS dataset originally collected as part of a randomized controlled trial of an online skills training program. Participants were 78 mental health professionals, including clinical psychology graduate trainees and licensed clinicians, recruited through psychotherapy email listservs and direct communication with psychology graduate programs.

\textbf{Procedure.} Mental health professionals watched each mock patient video and provided a response as if they were the treating therapist, their response typically ranged from 30s to 2 minutes. The parent training study administered the FIS task at baseline, post-training, and 2-week follow-up, and we used all available response clips collected at these three time points, resulting in 12–18 clips per participant. To improve measurement stability, the same patient scenario could be administered again after an interval of 1–2 months, for up to three administrations. Each participant responded to 4-8 mock patient scenarios. According to the raw feature inventory, the original collection contains 890 therapist-side recordings from 78 therapist identities across eight patient scenarios. All participants provided consent for their recordings to be used in the development of an AI-based FIS feedback system for clinicians. The recorded response clips were rated independently by three trained research assistants. Raters were trained to reliability with expert ratings (\textit{ICC} > 0.80 on six consecutive clips) and attended weekly calibration meetings to prevent rater drift. The Interrater reliability ranged from 0.70 to 0.86 (\textit{ICC3k}\textsubscript{3k}), with 0.86 for the total score of the FIS. We used the average of the three ratings for each subscale in subsequent analyses.

\textbf{Annotation.} Each therapist-response video was independently rated by three trained FIS observer raters following the official \textit{Facilitative Interpersonal Skills Task and Rating Manual} \cite{anderson2019facilitative}. The eight 1--5 subscales are Emotional Expression; Empathy; Hope and Positive Expectations; Persuasiveness; Verbal Fluency; Alliance Rupture-Repair Responsiveness; Alliance Bond Capacity; and Warmth, Acceptance, and Understanding. We first average the three raters for each subscale and then compute the overall FIS score as the arithmetic mean of the eight averaged subscales, yielding nine prediction targets.

    
  


\subsection{Data Preprocessing}
\noindent  In accordance with ethical requirements and participant privacy protection, we converted the original recordings into de-identified feature representations in three streams: face, audio, and text. This strategy preserves behaviorally meaningful signals for automatic FIS scoring while removing directly identifiable facial and vocal information.

\textbf{Facial cues.} The visual stream is represented by structured facial behavior descriptors rather than raw face images. We use OpenFace 3.0\cite{hu2025openface} to extract complementary signals, including AU occurrence and intensity, 98-point facial landmarks, gaze directions, and detection confidence. We also retained timestamps and applied temporal smoothing to support temporal alignment and modeling.

\textbf{Audio cues.} The audio stream is converted into abstract acoustic representations. We extracted acoustic features with librosa\cite{mcfee2015librosa}, including 80-dimensional log-Mel spectrograms and 13-dimensional MFCCs\cite{mfcc}, and contextualized speech representations with a pre-trained Wav2Vec 2.0 model\cite{baevski2020wav2vec2}, which yields 768-dimensional sequence features. In addition, we computed segment-level prosodic descriptors, including pitch statistics, energy variation, pause patterns, spectral characteristics, and voice-quality-related measures. Together, these features capture affective delivery and speaking styles while avoiding direct exposure of speaker identity.

\textbf{Text cues.} For the language stream, speech is converted into transcripts and contextualized text representations. A Whisper-based\cite{bain2023whisperx} forced-alignment process provides word-level timestamps and transcripts, which enables synchronization with facial and acoustic features. The transcripts are then encoded with  \texttt{microsoft/deberta-v3-large}\cite{he2023debertav3} to obtain contextual token-level embeddings for downstream learning. In this way, the released text stream preserves the semantic content and temporal organization of therapist responses without exposing raw audiovisual identity cues.

Overall, the released dataset is fully de-identified with respect to personal vocal and facial identity. It consists of facial behavior descriptors, acoustic features and textual representations derived from therapist speech.

\subsection{Statistic and Analysis}

\begin{figure*}[htbp]
  \centering
  \begin{subfigure}[b]{0.48\linewidth}
    \centering
    \includegraphics[width=\textwidth]{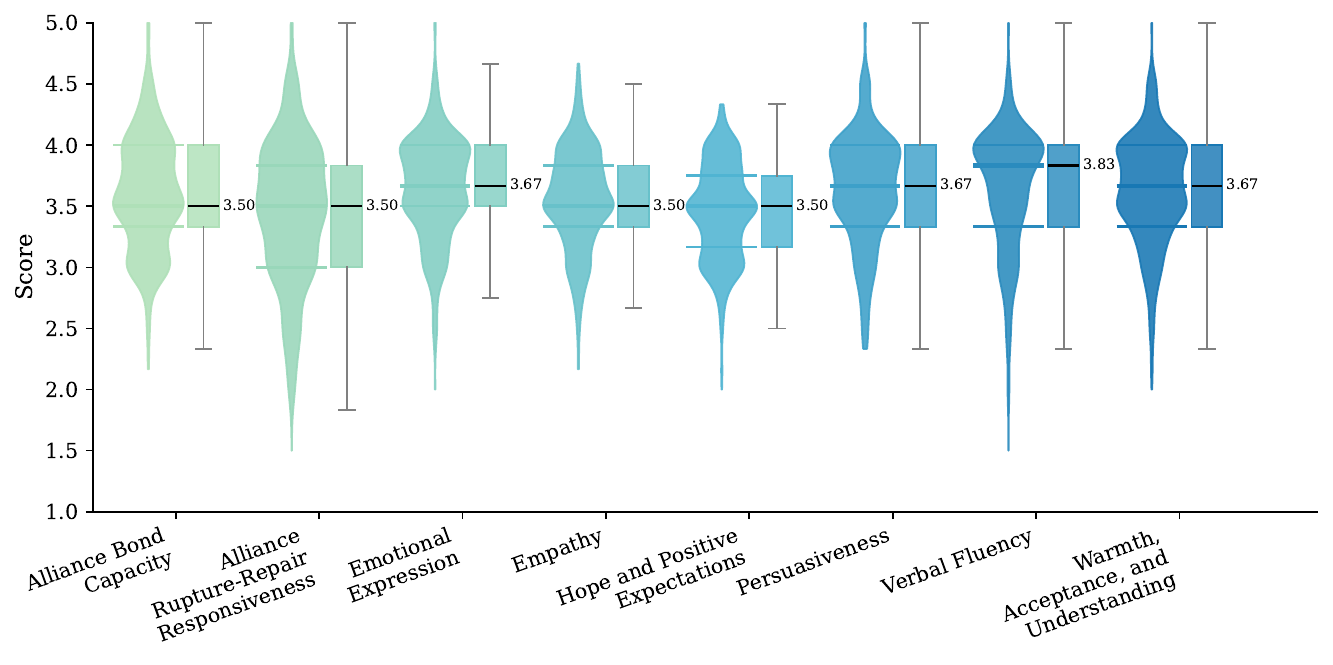}
    \caption{Label distribution of Therapy-FIS dataset.}
    \label{fig:stat_subfig_a}
  \end{subfigure}
  \hfill
  \begin{subfigure}[b]{0.48\linewidth}
    \centering
    \includegraphics[width=\textwidth]{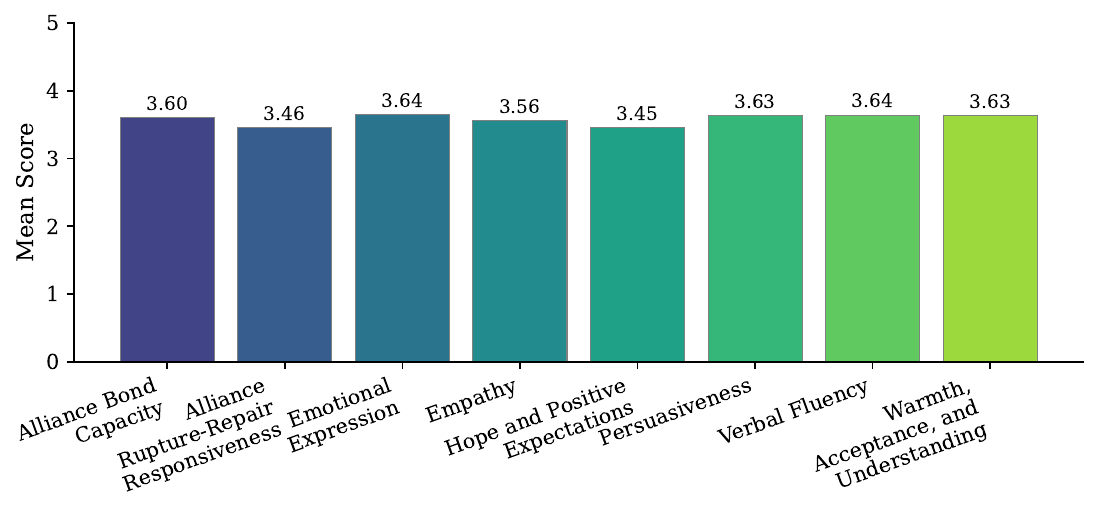}
    \caption{Mean score of Therapy-FIS dataset.}
    \label{fig:stat_subfig_b}
  \end{subfigure}
  \caption{Statistic and analysis of Therapy-FIS dataset. Panel(a) shows the score distribution for each label in Therapy-FIS, and panel(b) shows the mean scores of therapists under each label.}
  \label{fig:stat_analysis}
\end{figure*}

Therapy-FIS contains 890 therapist clips from 78 mental health professionals together with 8 mock patient recordings, totaling 16 hours and 51 minutes of interaction data. The data statistics are summarized in Table~\ref{tab:dataset_stats}, therapist clips have an average duration of 67.29~s and reach up to 278.19~s, whereas mock patient recordings are longer on average (96.82~s). Figure~\ref{fig:stat_analysis} further shows that the score distribution is centered in the mid-to-high range rather than collapsing to a narrow extreme. The overall FIS score averages 3.58 on a scale of 1--5, with most samples concentrated in the 3.5--4.0 interval. Among the subscales, emotional expression, verbal fluency, and persuasiveness have the highest mean scores, while alliance rupture-repair responsiveness and hope/positive expectation are relatively lower, suggesting that relational repair and future-oriented support remain comparatively challenging dimensions. Together with the temporally aligned visual, acoustic, and textual representations summarized in Table~\ref{tab:dataset_stats}, this score distribution indicates that Therapy-FIS provides both rich behavioral evidence and sufficient label variation for continuous FIS prediction.

\begin{table}[htbp]
  \centering
  \small
  \begin{tabular*}{\columnwidth}{@{\extracolsep{\fill}}lccc@{}}
    \toprule
    \textbf{Source} & \textbf{Counts} & \textbf{Mean (s)} & \textbf{Max (s)} \\
    \midrule
    Therapist & 890 & 67.29 & 278.19 \\
    Patient & 8 & 96.82 & 123.66 \\
    \bottomrule
  \end{tabular*}
  \vspace{0.45em}
  \begin{tabular*}{\columnwidth}{@{\extracolsep{\fill}}lp{0.5\columnwidth}r@{}}

    \textbf{Modality} & \textbf{Feature} & \textbf{Shape} \\
    \midrule
    \multirow{4}{*}{Face} & Action units & $T{\times}36$ \\
     & Facial landmarks (98 points, $xy$) & $T{\times}196$ \\
     & Gaze direction (2-D) & $T{\times}2$ \\
     & Detection confidence & $T{\times}2$ \\
    \midrule
    \multirow{4}{*}{Audio} & Wav2Vec~2.0 frame embeddings & $T{\times}768$ \\
     & Prosody (pitch, energy, pauses, spectral stats) & $T{\times}15$ \\
     & Log-Mel spectrogram  & $T{\times}80$ \\
     & MFCCs  & $T{\times}13$\\
    \midrule
    \multirow{2}{*}{Text} & DeBERTa-v3-large contextual token embeddings & $N{\times}1024$ \\
     & Transformed context  &  $N$ \\
    \bottomrule
  \end{tabular*}
  \normalsize
  \caption{Therapy-FIS scale and feature layout. $T$ denotes time steps aligned to OpenFace frames; $N$ is the number of transcript tokens.}
  \label{tab:dataset_stats}
\end{table}

\subsection{Ethics Review and License}

Before data collection, all participants were informed of the study procedures, including the recording process and the intended research use of the collected data. Written informed consent was obtained from all participants for conversation recording and for the use of the resulting data in scientific publications, including conference and journal papers. To protect participant privacy, the released dataset does not include raw audiovisual recordings and instead provides de-identified representations only.

The dataset is released under the CC BY-NC-SA 4.0 license. This license allows use, distribution, and adaptation of the dataset for non-commercial purposes, provided that appropriate attribution is given and any derivative works are distributed under the same license. Commercial use is not permitted.

\section{Benchmark}
We aim to establish a practical benchmark for automatic, privacy-preserving FIS assessment from derived behavioral features of standardized FIS responses. The benchmark tests whether FIS scores can be reliably predicted from facial, vocal, and language representations under the public release protocol.

\subsection{Tasks and Evaluation Protocol}
Each benchmark sample contains synchronized features extracted from a therapist or patient clip. The visual stream is represented by OpenFace-based facial behavior descriptors, including action units, facial landmarks, gaze directions, and confidence scores. The speech stream includes contextualized Wav2Vec 2.0 representations. The language stream is represented by token-level embeddings from DeBERTa-v3-large. Dataset timestamps make it possible to align text with facial and acoustic behavior on a shared timeline. The prediction target is a 9-dimensional continuous FIS vector consisting of eight subscales and their overall FIS score.

We study automatic FIS assessment under two complementary settings. \textbf{Task 1 (therapist-only FIS assessment)} scores the therapist solely from therapist-side features. \textbf{Task 2 (therapist--patient FIS assessment)} includes both therapist and mock-patient streams from the same therapy scenario, allowing the model to capture therapist--patient contingency and test whether dyadic modeling improves therapist FIS prediction.

\textbf{Evaluation Metrics.} We treat both tasks as 9-dimensional regression. For target $d\in\{1,\ldots,9\}$, let $\mu_d,\sigma_d^2$ and $\hat{\mu}_d,\hat{\sigma}_d^2$ denote the means and variances of the human ratings and predictions, and let $\sigma_{d\hat d}$ denote their covariance. We define
\begin{equation}
\begin{aligned}
  D_d
  &=\sigma_d^2+\hat{\sigma}_d^2+(\mu_d-\hat{\mu}_d)^2,\\
  \mathrm{CCC}_d
  &=\frac{2\sigma_{d\hat d}}{D_d},\\
  \mathrm{MacroCCC}
  &=\frac{1}{9}\sum_{d=1}^{9}\mathrm{CCC}_d.
\end{aligned}
\label{eq:macro_ccc}
\end{equation}
MSE is likewise averaged over all samples and all nine targets. FIS assessment prioritizes systematic agreement with expert observer ratings---capturing correlation, bias, and scale concordance---rather than minimizing pointwise error alone; we therefore treat macro CCC as the primary metric and report MSE as a complementary measure of absolute error.

\textbf{Experimental Setup.} We use 5-fold GroupKFold with grouping by therapist identity (cross-subject): each fold holds out a disjoint therapist group for testing, and the remaining therapists are split into training and validation. All methods predict the same 8-dimensional continuous FIS target. In Task~1, the input contains therapist-side features only. In Task~2, the input contains both therapist and patient feature streams from the same therapy scenario, while the supervision target remains the therapist FIS scores. We report mean~$\pm$~std over the five folds.

\textbf{Training Details.} All benchmark models are optimized with AdamW and trained for 40 epochs with learning rate $3\times10^{-4}$ and weight decay $3\times10^{-2}$; Task~2 uses a smaller batch size because of the larger memory footprint of dyadic modeling.

\subsection{DyadicFIS Baseline}

\begin{figure*}
    \includegraphics[width=0.9\textwidth]{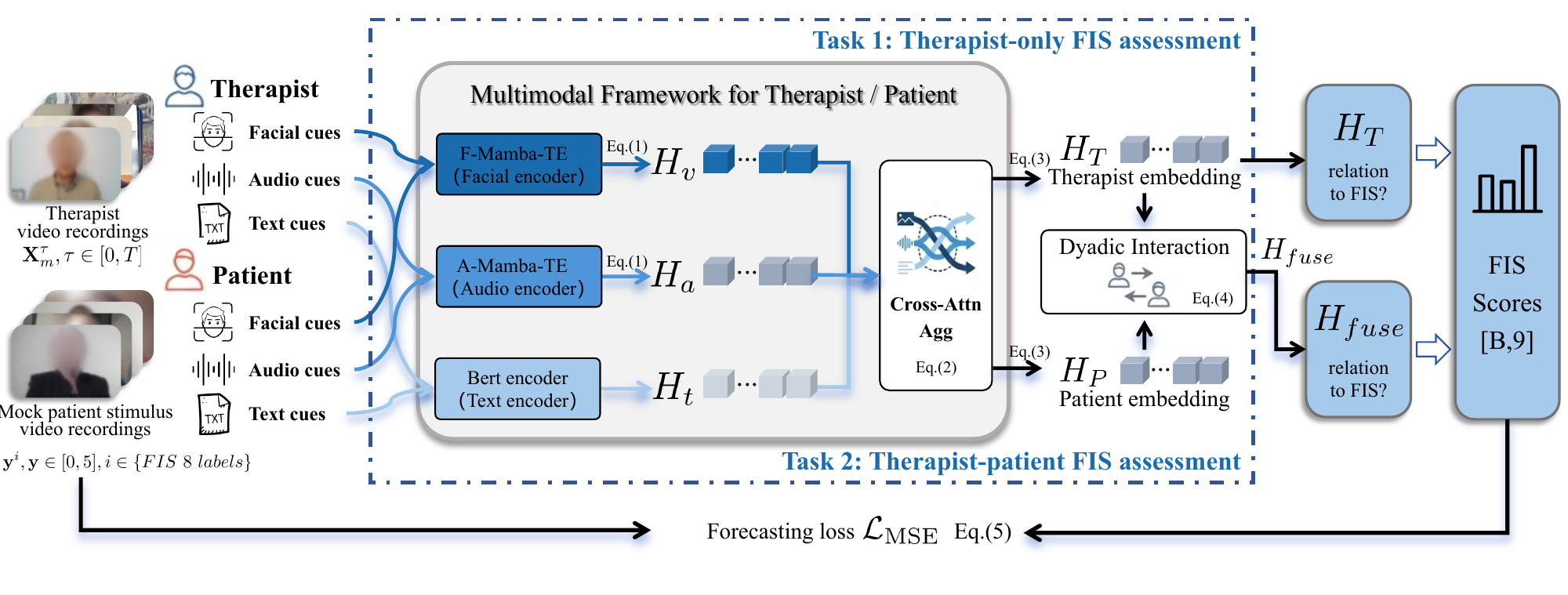}
    \caption{Architecture of DyadicFIS for Task~1 and Task~2. Inputs are facial, audio, and textual features. Task~1 uses therapist-only \{M\}-Mamba-TE and Cross-Attn-Agg; Task~2 additionally applies Dyadic Interaction over therapist and patient embeddings. Both tasks predict FIS scores.}
    \label{figure:model}
\end{figure*}

To address the two tasks, we propose DyadicFIS. For Task~1, therapist facial/acoustic/textual streams are encoded by \{M\}-Mamba-TE, fused by Cross-Attn-Agg, and mapped to FIS scores. For Task~2, the same pipeline is applied to therapist and mock-patient inputs, followed by a Dyadic Interaction block.

\textbf{\{M\}-Mamba-TE}:
For each modality $m \in \{v, a, t\}$, let $\mathbf{X}_m$ be the input sequence. As in Figure~\ref{figure:model}, facial and audio cues use F-Mamba-TE and A-Mamba-TE, while text uses a contextual text encoder (Bert encoder in the figure; DeBERTa embeddings in practice). Each stream is projected by $W_m$, compressed by latent-query aggregation $\mathrm{Agg}(\mathbf{Q}_m,\cdot)$, and temporally modeled:
\begin{equation}
  \mathbf{H}_m=\mathrm{Mamba}_m\!\left(\mathrm{Agg}\!\left(\mathbf{Q}_m, W_m\mathbf{X}_m\right)\right).
\end{equation}
For text, $\mathrm{Mamba}_t$ reduces to a lightweight feed-forward block because the embeddings are already contextualized.

\textbf{Cross-Attn-Agg}:
Let $\mathbf{H}_{\mathrm{all}}=[\mathbf{H}_v;\mathbf{H}_a;\mathbf{H}_t]$. Cross-Attn-Agg injects complementary evidence with gated cross-attention:
\begin{equation}
\begin{aligned}
  \mathbf{A}_m &= \mathrm{Attn}(\mathbf{H}_m,\mathbf{H}_{\mathrm{all}},\mathbf{H}_{\mathrm{all}}), \\
  \mathbf{g}_m &= \sigma\!\left(W_g[\mathbf{H}_m;\mathbf{A}_m]\right), \\
  \tilde{\mathbf{H}}_m &= \mathrm{LN}\!\left(\mathbf{H}_m+\mathbf{g}_m\odot\mathbf{A}_m\right).
\end{aligned}
\end{equation}
A modality-level gate reweights pooled modality summaries, and the visual stream serves as the temporal reference to form therapist embedding $\mathbf{H}_{\mathrm{T}}$ (and $\mathbf{H}_{\mathrm{P}}$ in Task~2) via length-preserving merge attention:
\begin{equation}
  \mathbf{H}_{\mathrm{T}} \leftarrow \mathrm{LN}\!\left(\mathbf{H}_{\mathrm{T}}+\mathbf{g}_{\mathrm{merge}}\odot\mathrm{Attn}(\mathbf{H}_{\mathrm{T}},\tilde{\mathbf{H}}_{m},\tilde{\mathbf{H}}_{m})\right),
\end{equation}
initialized by $\mathbf{H}_{\mathrm{T}}=\tilde{\mathbf{H}}_{v}$ for $m\in\{a,t\}$. A routed multi-view readout over $\mathbf{H}_{\mathrm{T}}$ then yields the Task~1 role representation.

\textbf{Dyadic Interaction}:
In Task~2, \{M\}-Mamba-TE and Cross-Attn-Agg produce $\mathbf{H}_{\mathrm{T}}$ and $\mathbf{H}_{\mathrm{P}}$. Dyadic Interaction exchanges context on the compressed sequences:
\begin{equation}
\begin{aligned}
  \mathbf{H}_{\mathrm{T}}' &= \mathrm{FFN}\!\left(\mathrm{LN}\!\left(\mathbf{H}_{\mathrm{T}}+\mathbf{g}_{\mathrm{T}}\odot\mathrm{Attn}(\mathbf{H}_{\mathrm{T}},\mathbf{H}_{\mathrm{P}},\mathbf{H}_{\mathrm{P}})\right)\right), \\
  \mathbf{H}_{\mathrm{P}}' &= \mathrm{FFN}\!\left(\mathrm{LN}\!\left(\mathbf{H}_{\mathrm{P}}+\mathbf{g}_{\mathrm{P}}\odot\mathrm{Attn}(\mathbf{H}_{\mathrm{P}},\mathbf{H}_{\mathrm{T}},\mathbf{H}_{\mathrm{T}})\right)\right), \\
  \mathbf{H}_{\mathrm{fuse}} &= [\mathbf{z}_{\mathrm{T}};\mathbf{z}_{\mathrm{P}};\mathbf{z}_{\mathrm{T}}-\mathbf{z}_{\mathrm{P}}],
\end{aligned}
\end{equation}
where $\mathbf{z}_{\mathrm{T}}$ and $\mathbf{z}_{\mathrm{P}}$ are pooled interacted states, and an MLP maps $\mathbf{H}_{\mathrm{fuse}}$ to FIS scores.

We optimize the forecasting loss
\begin{equation}
  \mathcal{L}=\mathcal{L}_{\mathrm{MSE}}+\lambda\,\mathcal{L}_{\mathrm{CCC}},
\end{equation}
with $\mathcal{L}_{\mathrm{MSE}}=\frac{1}{N}\sum_{i=1}^{N}\|\hat{\mathbf{y}}_i-\mathbf{y}_i\|_2^2$ and $\lambda=0.5$.

Besides DyadicFIS, we benchmark temporal backbones---TimeSformer\cite{2021timsformer}, DLinear\cite{2023transformers}, KANAD\cite{25KANAD}, Reformer\cite{2020reformer}, and MSGNet\cite{2024msgnet}---and multimodal fusion methods adapted to FIS regression, including JMT\cite{waligora2024jmt}, MDF\cite{jian2026mdf}, and CLGSI\cite{yang-etal-2024-clgsi}.


\subsection{Main Results}
Table~\ref{tab:baseline-results} compares DyadicFIS with temporal and fusion baselines. DyadicFIS achieves the highest macro CCC on Task~1 ($0.2807\pm0.0743$), surpassing the strongest external baseline, MDF, by $0.0607$. On Task~2, it achieves $0.3299\pm0.0556$, exceeding MDF by $0.0364$. DyadicFIS does not yield the lowest MSE on either task ($0.2566$ and $0.2186$, respectively), MDF instead achieves the lowest pointwise error. This difference in ranking reflects the distinct properties of the two metrics: CCC measures correlation together with bias and scale agreement, whereas MSE measures residual magnitude. The Task~2 result is numerically higher than Task~1, suggesting that modeling therapist--mock-patient contingency is beneficial in this standardized FIS setting.

\subsection{Component and Modality Ablation}
 \begin{table}[htbp]
\centering
\scriptsize
\setlength{\tabcolsep}{3.5pt}
\begin{tabular}{lccc}
\toprule
Task & Model & MSE$\downarrow$ & CCC$\uparrow$ \\
\midrule
\multirow{9}{*}{T1} & TimeSformer(2021)   & 0.2027$\pm$0.0357 & 0.2125$\pm$0.0831 \\
                    & DLinear(2023)       & 0.2256$\pm$0.0283 & 0.0961$\pm$0.0408 \\
                    & Reformer(2020)      & 0.2211$\pm$0.0352 & 0.1807$\pm$0.0400 \\
                    & MSGNet(2024)        & 0.2184$\pm$0.0274 & 0.1759$\pm$0.0260 \\
                    & KANAD(2025)         & 0.2096$\pm$0.0553 & 0.2043$\pm$0.0434 \\
                    & JMT(2024)           & 0.2156$\pm$0.0416 & 0.1286$\pm$0.0316 \\
                    & MDF(2026)           & \textbf{0.1858$\pm$0.0333} & 0.2200$\pm$0.1389 \\
                    & CLGSI(2024)         & 0.2201$\pm$0.0371 & 0.1376$\pm$0.1107 \\
                    & DyadicFIS(ours)     & 0.2566$\pm$0.0420 & \textbf{0.2807$\pm$0.0743} \\
\midrule
\multirow{9}{*}{T2}
                    & TimeSformer(2021)   & 0.1896$\pm$0.0287 & 0.2651$\pm$0.0313 \\
                    & DLinear(2023)       & 0.2245$\pm$0.0223 & 0.1878$\pm$0.0346 \\
                    & Reformer(2020)      & 0.2151$\pm$0.0499 & 0.1950$\pm$0.0460 \\
                    & MSGNet(2024)        & 0.2237$\pm$0.0536 & 0.1925$\pm$0.0367 \\
                    & KANAD(2025)         & 0.2047$\pm$0.0486 & 0.1845$\pm$0.0413 \\
                    & JMT(2024)           & 0.2005$\pm$0.0518 & 0.1641$\pm$0.0674 \\
                    & MDF(2026)           & \textbf{0.1820$\pm$0.0398} & 0.2934$\pm$0.0742 \\
                    & CLGSI(2024)         & 0.2042$\pm$0.0299 & 0.1973$\pm$0.0801 \\
                    & DyadicFIS(ours)     & 0.2186$\pm$0.0339 & \textbf{0.3299$\pm$0.0556} \\
\bottomrule
\end{tabular}
\normalsize
\caption{5-fold cross-subject results (mean$\pm$std) for Task~1 and Task~2.}
\label{tab:baseline-results}
\end{table}
Because Task~2 reflects the complete FIS measurement pipeline, we use this setting to isolate the contribution of each component of DyadicFIS and assess the stand-alone sufficiency of each modality. Table~\ref{tab:ablation-task2} reports both analyses. All variants use the same therapist-disjoint splits, training protocol.

\begin{table}[htbp]
\centering
\small
\setlength{\tabcolsep}{4pt}
\renewcommand{\arraystretch}{0.95}
\begin{tabular}{@{}llc@{}}
\toprule
Category & Variant & CCC$\uparrow$ \\
\midrule
Reference & Full & \textbf{0.3299$\pm$0.0556} \\
\midrule
\multirow{3}{*}{Module}
 & w/o centroid residual  & 0.3076$\pm$0.0689 \\
 & w/o dyadic interaction & 0.3234$\pm$0.0556 \\
 & w/o modality gate      & 0.3213$\pm$0.0907 \\
\midrule
\multirow{3}{*}{Modality only}
 & Audio only & 0.3241$\pm$0.0744 \\
 & Face only  & 0.0105$\pm$0.0829 \\
 & Text only  & 0.2807$\pm$0.0541 \\
\bottomrule
\end{tabular}
\caption{Task~2 ablations under the therapist-disjoint 5-fold protocol (mean$\pm$std). Each module variant removes one component from Full, whereas each modality-only variant retains the named modality for both roles.}
\label{tab:ablation-task2}
\end{table}

Removing any of the three components lowers the primary CCC metric. Removing the centroid residual produces the largest decrease ($\Delta\mathrm{CCC}=-0.0223$) and increases MSE by $0.0110$. Removing the modality gate lowers CCC by $0.0086$ and increases MSE by $0.0282$. Removing dyadic interaction lowers CCC by $0.0064$ but reduces MSE by $0.0090$. Thus, dyadic interaction improves rating concordance, the primary criterion, but not pointwise accuracy.

All modality-only variants perform below multimodal Full in CCC. Audio only is the strongest, reaching $0.3241$---only $0.0058$ below Full---with a similarly small MSE difference. Text only retains moderate concordance ($0.2807$), whereas Face only falls to $0.0105$. These variants measure stand-alone sufficiency rather than the conditional contribution of each modality. The results indicate that the acoustic representation contains most of the independently predictive signal, but they do not show that the other modalities are unnecessary in multimodal fusion.

\section{Discussion and Limitations}
Taken together, the results support a contextual view of FIS assessment. FIS scores reflect how appropriately a therapist responds to a preceding stimulus, rather than the therapist's behavior in isolation. Therapist language, vocal behavior, and facial expression characterize the response, while mock-patient cues establish the interpersonal and clinical context against which that response is judged. Task~1 therefore estimates how much information is recoverable from the response alone. Task~2 more closely mirrors the rating procedure by conditioning prediction on both sides of the exchange. Its higher CCC is consistent with the value of contextual information. 

This contextual interpretation also clarifies the apparent disagreement between the evaluation metrics. DyadicFIS ranks first by CCC on both tasks, whereas MDF obtains lower MSE. Since most labels lie in the mid-to-high range, predictions concentrated near the distributional center can achieve relatively small squared error while presenting less differences among responses. By contrast, CCC penalizes errors in correlation, location, and scale, which makes it more aligned with the goal of reproducing variation in observer ratings. One plausible account is that MDF behaves more conservatively while DyadicFIS better preserves between-sample variation. The ranking should therefore be understood as evidence that model selection depends on the intended use of the score. CCC is appropriate when concordance with trained raters is central, while MSE remains important when the magnitude of individual prediction errors carries practical consequences. Future evaluation should report calibration, subscale-level performance, confidence intervals for pairwise differences, and uncertainty estimates in addition to aggregate CCC and MSE.

The Task~2 ablations refine this interpretation by showing where predictive information may arise. Removing the centroid residual, modality gate, or dyadic interaction lowers mean CCC, which supports retaining all three components under the primary criterion. At the same time, the small change produced by the dyadic block and its opposite effect on MSE indicate that this component improves concordance rather than regression quality in every sense. The modality-only results require similar care. Audio alone approaches the full model, text retains moderate concordance, and facial descriptors alone perform poorly. This pattern may reflect the importance of prosody, pacing, and vocal affect in the standardized responses. It may also reflect differences in feature quality, temporal resolution, missingness, and signal-to-noise ratio across the released streams. 

Two limitations concern the labels and the release format. The labels average three trained observers' ratings of standardized role-play. Although prior FIS research links performance-based scores to therapeutic alliance and patient outcomes \cite{anderson_therapist_2009, anderson_therapist_2016}, these ratings are not direct measurements of routine clinical sessions, and averaging hides rater disagreement. Releasing de-identified features rather than raw recordings reduces identity exposure, but fixes the upstream extractors and limits end-to-end training and independent error analysis. In future work, we will retain rater-level uncertainty, validate predictions against clinical behavior and outcomes, and explore privacy-audited controlled access under appropriate consent and institutional review.

\section{Conclusion}
This work presents a dataset and benchmark for FIS assessment in psychotherapeutic settings, together with DyadicFIS, a baseline for automatic FIS prediction. Under a therapist-disjoint 5-fold protocol, DyadicFIS achieves the strongest concordance with human ratings (CCC) on both therapist-only and therapist--mock-patient tasks, providing a reproducible starting point for scalable evaluation of therapeutic interpersonal skills.

\clearpage
\bibliography{references}

\end{document}